# Temporal Matrix Completion with Locally Linear Latent Factors for Medical Applications

Frodo KS Chan[1], Andy J Ma[1], Pong C Yuen[1], Terry CF Yip[2], Yee-Kit Tse[2], Vincent WS Wong[2], Grace LH Wong[2]

[1]Hong Kong Baptist University
[2]The Chinese University of Hong Kong
{frodochan; andyjhma; pcyuen}@comp.hkbu.edu.hk
terryfungyip@gmail.com, {yktse; wongv; wonglaihung}@cuhk.edu.hk

## Abstract

Regular medical records are useful for medical practitioners to analyze and monitor patient's health status especially for those with chronic disease, but such records are usually incomplete due to unpunctuality and absence of patients. In order to resolve the missing data problem over time, tensor-based model is suggested for missing data imputation in recent papers because this approach makes use of low rank tensor assumption for highly correlated data. However, when the time intervals between records are long, the data correlation is not high along temporal direction and such assumption is not valid. To address this problem, we propose to decompose a matrix with missing data into its latent factors. Then, the locally linear constraint is imposed on these factors for matrix completion in this paper. By using a publicly available dataset and two medical datasets collected from hospital, experimental results show that the proposed algorithm achieves the best performance by comparing with the existing methods.

## Introduction

Regular medical records, which contain patients' medical examinations, diagnosis, treatments, etc. across time, are important sources of assessing the patient's health status. Based on such assessment, the medical practitioners can effectively monitor disease progress and treatment response and provide valuable advices and proper treatments to patients. Furthermore, some patients with chronic disease can develop a fatal illness without proper guidance and monitoring. For example, patients with non-alcoholic fatty liver disease are at risk of cirrhosis and hepatocellular carcinoma without proper management (Wong et al. 2012).

Although regular medical records are beneficial for the patients, missing data is an intractable problem which may hamper precise decisions by medical practitioners. Such problem happens because of many reasons from patients, e.g. forgetful character, busy work, fear of certain tests etc. Even though self-report quality-of-life-questionnaires, which can be used to assess how the patients are affected over time by a disease, are employed, the missing data is common and unavoidable (Garg et al. 2014).

To tackle the missing data problem, one of existing popular methods is mean imputation, but it does not consider the relationship among attributes. In recent years, there are many research works on temporal matrix completion, which is applicable on compressed sensing (Chen and Chi 2014), seismic data reconstruction (Yang, Ma, and Stanley 2014; Yuan and Zhan 2016), network data completion (Roughan 2011), traffic data completion (Tan et al. 2013; Asif et al. 2013), medical signal (e.g. EEG, dynamic MRI) reconstruction (Mardani, Mateos, and Giannakis 2013; Singh et al 2014; Majumdar, Gogna, and Ward 2014), temporal collaborative filtering (Koren 2009; Xiong et al. 2013), and visual data and video restoration (Liu et al. 2013; Wang, Nie, and Huang 2014; Kim, Sim, and Kim 2015). In those works, tensor-related approach is dominant on solving the temporal matrix completion problem and canonical polyadic (CP) tensor decomposition techniques are commonly used. Since tensor rank computation is NP-hard, Liu et al. (Liu et al. 2013) suggested new tensor trace norm for tensor completion on visual data. However, Wang et al. (Wang, Nie, and Huang 2014) pointed out that minimizing trace norm of a tensor has to fulfill an assumption that the video frames are highly correlated in the temporal axis. Such assumption may not be true when the video clip has multiple different episodes. In medical record imputation, Dauwels et al. (Dauwels et al. 2012) and Garg et al. (Garg et al. 2014) suggested to use CP tensor decomposition for imputing quality-of-life questionnaire, which is used to assess the quality of life of a patient, treatment and disease progress, and quality of care, recently. They argued that tensor decomposition can exploit correlations in the data. However, when the periods of some health checkups and follow-up consultations are long, the data correlation is not high along the temporal direction.

In this paper, a novel algorithm is proposed to impose the locally linear constraint on the latent factors of medical record for matrix completion. Since the time interval between records can be long, the data correlation assumption is not valid along the temporal direction for tensor decomposition and the latent factors of the medical record can change non-linearly over time. Thus, the proposed algorithm applies the locally linear constraint on the latent factors of the medical record for regularization and converts the tensor decomposition problem to matrix completion problem by using block matrix techniques. To solve the matrix completion problem, alternate least squares (ALS) approach is applied with Sylvester questions. The experimental results show that the proposed algorithm outperforms the existing works compared in experimental results.

The organization of this paper is as follows. The next section reviews some related works. In Section 3, the proposed algorithm is explained. Finally, Section 4 and 5 gives experimental results and conclusion.

## Related Works

The matrix completion problem is usually tackled by minimizing the norm of a matrix $Z \in \mathbb{R}^{m \times n}$ with missing data and the formulation is

$$\operatorname{argmin}_Z \operatorname{rank}(Z), \qquad (1)$$
$$\text{s.t. } Z_\Omega = M_\Omega,$$

where $\operatorname{rank}(Z)$ is the rank of $Z$ and the elements of $M$ in the set $\Omega$ are observed while others are missing. Since $\operatorname{rank}(\cdot)$ is non-convex function and difficult to optimize, the convex nuclear norm $\|Z\|_*$ is used to minimize the rank. Mazumber, Hastie, and Tibshirani (Mazumber, Hastie, and Tibshirani 2010) show that

$$\|Z\|_* = \min_{A,B} (\|A\|_F^2 + \|B\|_F^2), \qquad (2)$$

where $Z = AB'$. By using equation (1), Hastie et al. (Hastie et al 2015) introduced ALS method to solve the following equation

$$\min_{A,B} 0.5\|Z_\Omega - (AB')_\Omega\|_F^2 + \qquad (3)$$
$$0.5\lambda(\|A\|_F^2 + \|B\|_F^2).$$

The details of its algorithm can be referenced in (Hastie et al 2015). The intrisinic way is to apply these matrix completion methods on each matrix along temporal direction, but such method ignores the correlation between these matrices.

On the other hand, Dauwels et al. (Dauwels et al. 2012) and Garg et al. (Garg et al. 2014) suggested to use CP tensor decomposition for quality-of-life questionnaire imputation. In CP decomposition, a minimal sum of rank-one tensors forms a tensor $\mathcal{X} \in \mathbb{R}^{I_1 \times I_2 \times \ldots \times I_N}$, which is

$$\mathcal{X} = v_1 \circ v_2 \circ \ldots \circ v_N, \qquad (4)$$

where $v_m \in \mathbb{R}^{I_m \times R}$, and $v_i \circ v_j$ denotes the outer product of vectors $v_i$ and $v_j$. Then, the reconstruction error is minimized as

$$\min_{v_1, v_2, \ldots, v_N} 0.5\|\Lambda - v_1 \circ v_2 \circ \ldots \circ v_N\|_F^2 \qquad (5)$$
$$\text{s.t. } \Lambda_\Omega = T_\Omega,$$

where $\|\cdot\|_F$ is a Frobenius norm and $\Lambda$, $T$ are the $N$-mode tensors with identical size in each mode and the elements of $T$ in the set $\Omega$ are observed while others are missing. Although CP decomposition can break down the tensor into several individual components, the rank-1 tensor assumption is useful on highly correlated data only. When the time interval between medical records is long, this assumption is not reasonable for the medical records along temporal directon.

Since rank-1 tensor assumption is not always true, Liu et al. (Liu et al. 2013) considered to define the new trace norm, which is

$$\|\Lambda\|_* = \sum_{i=1}^N \alpha_i \|\Lambda_{(i)}\|_*, \qquad (6)$$

where $\Lambda_{(i)} \in \mathbb{R}^{I_i \times I_1 \ldots I_{i-1} I_{i+1} \ldots I_N}$ is the $i$-th mode on the tensor $\Lambda$ to unfold a tensor into a matrix. By using equation (6), the optimization algorithm becomes

$$\min_\Lambda \sum_{i=1}^N \alpha_i \|\Lambda_{(i)}\|_*, \qquad (7)$$
$$\text{s.t. } \Lambda_\Omega = T_\Omega,$$

By using equation (7), three different methods including SiLRTC, FaLRTC, and FaLRTC are developed based on different conditions. However, Wang, Nie, and Huang (Wang, Nie, and Huang 2014) pointed that their approach relies on highly correlated data in the temporal direction when their new tensor norm which is the average trace norms of tensor unfolded along each mode is minimized.

Instead, Wang, Nie, and Huang suggested to make use of temporal information between consecutive video frames by introducing a smoothness regularization and assume that the successive data in video should not change very much. They applied the temporally consistent constraint to regularize the norm minimization. However, such constraint is not applicable to medical records because the time interval between can be over a year. The main difference of their constraints and our proposed algorithm is that their consistency constraint was applied on whole video frame, but our proposed constraint is imposed on latent factors of the medical record, which cause the change of the medical record over time. Thus, our proposed constraint can effectively regularize the optimization on nonlinear changes of latent factors.

## Temporal Matrix Completion with Locally Linear Constraint

In order to impute the missing data, the proposed algorithm makes use of the locally linear constraint to regularize the norm minimization. Initially, the record with missing data is approximated by using two latent factors. Then, the locally linear constraints are imposed on the latent factors to form the optimization equation. To solve this equation, Frobenius norms are employed for rank minimization and block matrix techniques are applied to reformulate the equation. And then, the equation is solved by using ALS, which optimize one parameter by fixing the others at each time. After the latent factors are found from optimization, they are used to form final solution.

### Problem Formulation

In singular value decomposition (SVD), the matrix can be decomposed into three components ($U$, $D$, $V$). Since the medical record $F_t$ at time $t$ contains missing values and can be approximated by $\widetilde{F}_t = U_t D_t V_t'$, where $U_t$ and $V_t$ are latent factors on the patient-to-patient and item-to-item relationship. Note that item can be the questions from healthcare questionnaires, blood test results or health indicators in medical records. The decomposition on a medical record $F_t \sim \widetilde{F}_t$ at time $t$ can be illustrated in Fig. 1. To impute the missing value of medical records over time, the objective function is formed as

$$\min_{\widetilde{F}_t} 0.5 \sum_{t=1}^{T} \|W_t \odot (F_t - \widetilde{F}_t)\|_F^2 \quad (8)$$
$$+ \lambda \sum_{t=1}^{T} \|\widetilde{F}_t\|_*,$$

where $\|\cdot\|_*$ is a nuclear norm, $\odot$ is a Hadamard product, $\lambda$ is a scalar value and the element in $W_t$ is equal to 1 if the corresponding entry of $F_t$ is not missing and 0 otherwise. Furthermore, $\widetilde{F}_t$ can be decomposed as $O_t P_t'$, where $\widetilde{F}_t \in \mathbb{R}^{m \times n}$, $O_t \in \mathbb{R}^{m \times r}$, $P_t \in \mathbb{R}^{n \times r}$, $O_t = U_t \sqrt{D_t}$, and $P_t = V_t \sqrt{D_t}$. It implies that $O_t$ and $P_t$ are the latent factors of the patient-to-patient and item-to-item relationship at time $t$ because $\widetilde{F}_t \widetilde{F}_t' = U_t D_t^2 U_t' = O_t D_t O_t'$ and $\widetilde{F}_t' \widetilde{F}_t = V_t D_t^2 V_t' = P_t D_t P_t'$. By equation (2), it can be shown that

$\|\widetilde{F}_t\|_* = \min_{O_t, P_t} 0.5(\sum_{t=1}^{T} \|O_t\|_F^2 + \|P_t'\|_F^2)$ when $\widetilde{F}_t = O_t P_t'$. Then, the equation (8) becomes

$$\min_{O_t, P_t} 0.5 \sum_{t=1}^{T} \|W_t \odot (F_t - O_t P_t')\|_F^2 + \\ 0.5\lambda(\sum_{t=1}^{T} \|O_t\|_F^2 + \|P_t'\|_F^2). \quad (9)$$

In equation (9), $O_t$ and $P_t$ are the latent factors of the patient-to-patient and item-to-item correlation over time. By introducing the locally linear constraint on $O_t$ and $P_t$, the equation (9) becomes

$$\min_{O_t, P_t} 0.5 \sum_{t=1}^{T} \|W_t \odot (F_t - O_t P_t')\|_F^2 + \\ 0.5\lambda(\sum_{t=1}^{T} \|O_t\|_F^2 + \|P_t'\|_F^2) + \\ 0.5\gamma(\sum_{t=2}^{T-1} \|(O_{t+1} - O_t) - (O_t - O_{t-1})\|_F^2) + \\ 0.5\beta(\sum_{t=2}^{T-1} \|(P_{t+1} - P_t) - (P_t - P_{t-1})\|_F^2), \quad (10)$$

where $\gamma$ and $\beta$ are scalar values. Since the change of latent factors of medical records is nonlinear, it can be regularized by the locally linear constraints.

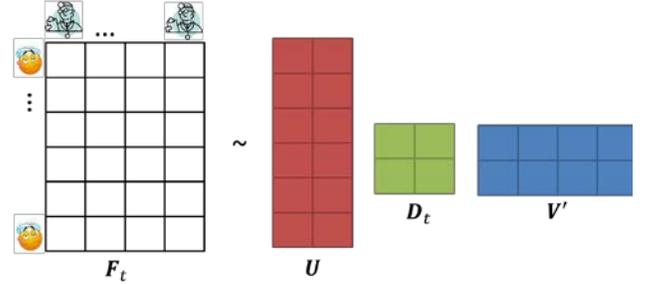

*Fig. 1 The illustration of the decomposition on a medical record at time $t$.*

### Optimization Algorithm

To effectively solve the equation (10), the technique of block matrix is employed for $t = (1, 2, \ldots, T)$. By defining $O = (O_1', \ldots, O_T')'$ and $P = (P_1', \ldots, P_T')'$, the matrix of medical records $F \in \mathbb{R}^{(T \times m) \times (T \times n)}$ is approximated by $OP'$, i.e. $F \sim OP'$. Then, $F$ and $W$ are defined as

$$F = \begin{bmatrix} F_1, F_2, \ldots, F_T \\ F_1, F_2, \ldots, F_T \\ \ldots \\ F_1, F_2, \ldots, F_T \end{bmatrix},$$

$$W = \begin{bmatrix} W_1, 0_{m \times n}, \ldots, 0_{m \times n} \\ 0_{m \times n}, W_2, \ldots, 0_{m \times n} \\ \ldots \\ 0_{m \times n}, 0_{m \times n}, \ldots, W_T \end{bmatrix}$$

where $0_{m \times n}$ is a $m \times n$ zero matrix. In $F$, the relationship $F_t \sim O_t P_t'$ can be maintained for each diagonal block of $F$. For the non-diagonal blocks of $F$, the assigned values are only for initialization and $W$ is used to mask out these blocks in calculating the convergence rate. In addition, to solve the time regularization problem, there are two matrices $Q_O \in \mathbb{R}^{((T-2)\times m)\times(T\times m)}$ and $Q_P \in \mathbb{R}^{((T-2)\times n)\times(T\times n)}$ defined as

$$\begin{bmatrix} I_{m\times m}, -2I_{m\times m}, I_{m\times m}, 0_{m\times m}, \ldots, 0_{m\times m} \\ 0_{m\times m}, \ldots, I_{m\times m}, -2I_{m\times m}, I_{m\times m}, \ldots, 0_{m\times m} \\ 0_{m\times m}, \ldots, 0_{m\times m}, I_{m\times m}, -2I_{m\times m}, I_{m\times m} \end{bmatrix} \text{ and}$$

$$\begin{bmatrix} I_{n\times n}, -2I_{n\times n}, I_{n\times n}, 0_{n\times n}, \ldots, 0_{n\times n} \\ 0_{n\times n}, \ldots, I_{n\times n}, -2I_{n\times n}, I_{n\times n}, \ldots, 0_{n\times n} \\ 0_{n\times n}, \ldots, 0_{n\times n}, I_{n\times n}, -2I_{n\times n}, I_{n\times n} \end{bmatrix},$$

where $I_{m\times m}$, $I_{n\times n}$ are $m \times m$ and $n \times n$ identity matrix, and $0_{m\times m}$, $0_{n\times n}$ are $m \times m$ and $n \times n$ zero matrix. By using $F$, $W$, $Q_O$ and $Q_P$, the equation (10) is reformulated as

$$\min_{O,P} 0.5\|W\odot(F - OP')\|_F^2 + \\ 0.5\lambda(\|O\|_F^2 + \|P'\|_F^2) + \\ 0.5\alpha(\|Q_O O\|_F^2) + 0.5\beta(\|P'Q_P\|_F^2). \quad (11)$$

Although the last three terms of equation (11) is differentiable, the first term is still difficult to be resolved. By taking reference from (Hastie et al 2015), the first term is derived as

$$W\odot(F - OP') \\ = W\odot F - (1 - W)\odot(OP') - OP' \quad (12)$$

Suppose $O_{(i)}$ and $P_{(i)}$ have current estimates at iteration $i$, $P_{(i+1)}$ can be computed by fixing $O_{(i)}$ and replacing the first occurrence of $OP'$ with the current estimate, i.e. $\widehat{F} = W\odot F - (1 - W)\odot(O_{(i)}P_{(i)}')$. Then, the equation (11) becomes

$$\min_{\widetilde{P}} 0.5\|\widehat{F} - O_{(i)}P_{(i+1)}'\|_F^2 + \\ 0.5\lambda\left(\|O_{(i)}\|_F^2 + \|P_{(i+1)}'\|_F^2\right) + \\ 0.5\alpha(\|Q_O O_{(i)}\|_F^2) + 0.5\beta\left(\|P_{(i)}'Q_P\|_F^2\right). \quad (13)$$

From equation (13), new $P_{(i+1)}$ is obtained by solving

$$O_{(i)}'\widehat{F} = (\lambda I + O_{(i)}'O_{(i)})P_{(i+1)}' + \\ P_{(i+1)}'(\beta Q_P'Q_P), \quad (14)$$

where $I$ is the identity matrix. By fixing $P_{(i+1)}$ and replacing the first occurrence of $OP'$ in equation (12) with the current estimates, i.e. $\widehat{F}' = W'\odot F' - (1 - W')\odot(P_{(i+1)}O_{(i)}')$, $O_{(i+1)}$ is computed. Similarly, the equation (11) becomes

$$\min_{\widetilde{O}} 0.5\|\widehat{F}' - P_{(i+1)}O_{(i+1)}'\|_F^2 + \\ 0.5\lambda\left(\|P_{(i+1)}\|_F^2 + \|O_{(i+1)}'\|_F^2\right) + \\ 0.5\alpha(\|O_{(i+1)}'Q_O'\|_F^2) + 0.5\beta\left(\|P_{(i+1)}'Q_P\|_F^2\right). \quad (15)$$

From equation (15), new $\widetilde{O}$ is obtained by solving

$$P_{(i+1)}'\widehat{F}' = (\lambda I + P_{(i+1)}'P_{(i+1)})O_{(i+1)}' \quad (16) \\ + O_{(i+1)}'(\alpha Q_O'Q_O).$$

Equation (14) and (16) are Sylvester Equation $AX + XB = C$, which can be reformed as

$$(I\otimes A + B^T\otimes I)X(:) = C(:), \quad (17)$$

where $\otimes$ is a Kronecker tensor product, $I$ is the identity matrix, $X(:)$ and $C(:)$ represent $X$ and $C$ as single column vectors. Thus, the equation (14) and (16) can be solved effectively by reforming them as systems of linear equations (equation (17)). The overall algorithm is presented in Algorithm 1.

In the Algorithm 1, the initial value of $\widehat{U}$, $\widehat{D}$, and $\widehat{V}$ can be randomly generated. Then, $O$ and $P$ are assigned as $\widehat{U}\widehat{D}^{0.5}$ and $\widehat{V}\widehat{D}^{0.5}$. At each iteration, $O$ and $P$ are updated by solving the equation (14) and (16) and the convergence rate $C$ is calculated. The algorithm continues until the convergence rate exceeds a threshold or iteration limit is reached. In the implementation, the threshold value was $10^{-5}$. After the final values of $O$ and $P$ are obtained, $\widetilde{F} = OP'$ is calculated and its SVD is found as $UDV'$. In order to obtain the low rank representation of $\widetilde{F}$, some of diagonal values in $D$ are removed if they are less than threshold $\lambda$. The final result is $\widetilde{F} = US_\lambda(D)V'$.

---

Set $\lambda, \alpha, \beta, i = 1$. Initialize $\widehat{U}, \widehat{D}, \widehat{V}$,
$O_{(1)} = \widehat{U}\widehat{D}^{0.5}$,
$P_{(1)} = \widehat{V}\widehat{D}^{0.5}$,
initialize $\widehat{F}$ by filling missing values in $F$ as 0
**while** *not converaged* or *not exceeded iteration limit*
  1. Solve (14) to update $P_{(i+1)}$ by fixing $O_{(i)}$
  2. Update $\widehat{F}' = W'\odot F' - (1 - W')\odot(P_{(i+1)}O_{(i)}')$
  3. Solve (16) to update $O_{(i+1)}$ by fixing $P_{(i+1)}$
  4. Update $\widehat{F} = W\odot F - (1 - W)\odot(O_{(i+1)}P_{(i+1)}')$
  5. Compute the convergence rate $C$
    $= \|W\odot K_{new} - W\odot K_{old}\|_F^2 / \|W\odot K_{old}\|_F^2$,
    where $K_{new} = O_{(i+1)}P_{(i+1)}'$ and $K_{old} = O_{(i)}P_{(i)}'$
  6. $i = i + 1$
**end**
Obtain the new SVD of $K_{new} = UDV'$
Output the soft-threshold SVD $US_\lambda(D)V'$ where $S_\lambda$ is an operator to replace $D_{ii}$ with $\min(D_{ii} - \lambda, 0)$.

*Algorithm 1: Algorithm to solve equation (10)*

# Experimental Results

In order to evaluate the proposed algorithm in the previous section, mean imputation (meanImpute) method, CP decomposition (Tensor-CP) method (Dauwels et al. 2012; Garg et al. 2014), softImpute-ALS method (Hastie et al 2015), softImpute-SVD method (Mazumder, Hastie, and Tibshirani 2010), Low-Rank Tensor Completion (LRTC) method (Liu et al. 2013), and Low-Rank Tensor Completion with Spatio-Temporal Consistency (LRTC-STC) method (Wang, Nie, and Huang 2014) are employed for comparisons. Since meanImpute is one of the popular methods for data imputation, it is included for comparison. softImpute-ALS and softImpute-SVD are used for evaluation because they are two publicly available software for imputation in R. Since LRTC is related with tensor completion and relies on strong data correlation assumption, which is mentioned in (Wang, Nie, and Huang 2014), it can be compared with the proposed algorithm, which is not assumed to have strong data correlation along temporal direction. Besides, LRTC-STC is a recent study with consistency constraint, so it is compared with the proposed algorithm with locally linear constraint.

For the implementation, source of softImpute-ALS, softImpute-SVD, and LRTC are available. For Tensor-CT, Dauwels et al. and Garg et al. made use of CP decomposition, which can be obtained online (MATLAB Tensor Toolbox Version 2.5 2012). For LRTC-STC, it is implemented by the first author according to his best knowledge. In the experiments, softImpute-ALS and softImpute-SVD were performed with $\lambda = 0.25, 0.5, 1, 2, 4, 8$. Since they are general matrix completion method and they are designed for two dimensional matrix, they are applied on matrices along the temporal direction one by one. For Tesnor-CT, it is evaluated with $R = 1, 2, 3$. In LRTC, there are three different implementations, HaLRTC, FaLRTC, and SiLRTC, which are evaluated with default parameters except $\alpha = (1, 1, 1^{-6})$, $(1, 1, 1^{-3})$, $(1, 1, 1)$ for three methods and $\rho = 10^{-2}, 10^{-8}$ for HaLRTC. LRTC-STC is evaluated its internal parameters $p = 1.3, 1.5, 1.7$, $u = 1, 10^3, 10^6$ and $\alpha = 50$. In the experimental results, only the best result from different parameter settings is reported for each compared algorithm. For the proposed algorithm, the parameters are set as $\lambda = 4$, $\alpha = 10^{-3}$, $\beta = 10^{-3}$ for all experiments.

To compute the imputation error $E$, root-mean-square error (RMSE) is applied and the computed solution $F_{\text{sol}}$ and the ground truth $F$, i.e.

$$E = (\|W \odot (F_{\text{sol}} - F)\|_F^2 / N_{\text{miss}})^{0.5}, \qquad (17)$$

where $N_{\text{miss}}$ is the number of missing data of diagonal blocks. By using this measure, the best algorithm is the one with lowest RMSE.

In order to evaluate the performance of the imputation, three datasets are used. The first dataset is 2008-2010 beneficiary files from Data Entrepreneurs' Synthetic Public Use File (CMS 2014), which provides a realistic set of claims data in the public domain. These records contain the health conditions of beneficiaries and the records of 1000 beneficiaries with 14 attributes are randomly extracted to be the first dataset, which is called CMS dataset with 1000×14×3 dimensions in this paper, for evaluation. The 14 attributes are sex, race, and 12 health-related records (including the indicators of Alzheimer or related disorders or senile, heart failure, chronic kidney disease, cancer, chronic obstructive pulmonary disease, depression, ischemic heart disease, osteoporosis, rheumatoid arthritis, and diabetes). The second and third datasets, which are called HepB-I and HepB-II datasets in this paper, contain the medical records from patients who suffered from hepatitis B (HepB) for the evaluation of probable hepatocellular carcinoma (HCC) at the Prince of Wales Hospital, Hong Kong (Wong et al. 2013; Wong et al. 2014). 493 patient records from 2009-2011 and 326 patient records from 2008-2013 are extracted to form the HepB-I with 493×13×3 dimensions and HepB-II datasets with 326×11×6 dimensions. The attributes of the HepB-I datasets are sex and 12 blood test results (including tuberculosis (TB), alpha-fetoprotein (AFP), albumin, alkaline phosphatase (ALB), alkaline phosphatase (ALP), alanine aminotransferase (ALT), creatinine (ALT), international normalized ratio (INR), platelet (PLT), prothrombin time (PT), potassium (K), sodium (Na), and urea (Urea)) and the attributes of the HepB-II datasets are sex and 10 blood test results which are the same attributes of the blood test results in the HepB-I dataset except INR and PT.

To perform the experiments, 40%-90% observed data were chosen in each dataset and five datasets with different observed data were randomly generated for each percentage. Then, the imputation error of each algorithm was calculated by taking average of RMSEs from these five datasets. Each dataset was normalized to be zero mean and unit scaling before it was used for experiments. The corresponding experimental results are shown in Table 1-3.

In Table 1, the experimental results are reported based on the CMS dataset. In the 80%-90% observed data, the proposed algorithm performs worse than LRTC. But it obtains the best result at 40%-70% observed data and it outperforms the second best method at least 2% at 40%-50% observed data. In Table 2, HepB-I dataset is used for evaluation. From 40%-90% observed data, the proposed algorithm has the best performance. The proposed algorithm can outperform the second best at least 4.9% at 40%-50% observed data. In Table 3, the evaluation is performed wth HepB-II dataset. In this dataset, the proposed algorithm performs the best from 40%-90% observed data and it can be better than other methods at least 4 % at 40% observed data.

## Conclusion

Regular medical records, which systematically documents the patients' medical history and care across time, are useful for medical practitioners to analyze health status of patient and provide proper care and treatment. However, missing data problem is common in regular medical records.

In order to tackle the missing data problem in regular medical records, an algorithm, which imposes the locally linear constraints on the latent factors of the medical records, is proposed. Since the latent factors can change nonlinearly in time to cause the change of the medical records, the locally linear constraint can effectively regularize their nonlinear changes. In the experiments, the performance of the proposed algorithm is better than that of the comparison methods except at 80-90% observed data of the CMS dataset. The experimental results show the usefulness of locally linear constraint on the public dataset and two medical datasets.

Table 1: Imputation error on the CMS dataset with 40%-90% observed data

|  | 90% | 80% | 70% | 60% | 50% | 40% |
|---|---|---|---|---|---|---|
| meanImpute | 1.002 | 1.000 | 0.999 | 1.001 | 1.001 | 1.003 |
| Tensor-CP ($R=2$) | 0.898 | 0.891 | 0.942 | 0.989 | 1.684 | 1.427 |
| softImpute-ALS ($\lambda=8$) | 0.965 | 0.952 | 0.950 | 0.948 | 0.948 | 0.955 |
| softImpute-SVD ($\lambda=8$) | 0.965 | 0.951 | 0.950 | 0.948 | 0.948 | 0.956 |
| LRTC (FaLRTC) | **0.837** | **0.828** | 0.843 | 0.863 | 0.881 | 0.908 |
| LRTC-STC ($p=1.5$, $u=1$) | 0.939 | 0.930 | 0.934 | 0.939 | 0.942 | 0.954 |
| Proposed algorithm | 0.846 | 0.832 | **0.840** | **0.854** | **0.861** | **0.883** |

Table 2: Imputation error on the HepB-I dataset with 40%-90% observed data

|  | 90% | 80% | 70% | 60% | 50% | 40% |
|---|---|---|---|---|---|---|
| meanImpute | 0.936 | 1.002 | 0.995 | 0.994 | 0.991 | 1.000 |
| Tensor-CP ($R=2$) | 0.782 | 0.845 | 0.848 | 0.856 | 0.859 | 0.892 |
| SoftImpute-ALS ($\lambda=2$) | 0.784 | 0.851 | 0.855 | 0.878 | 0.891 | 0.925 |
| softImpute-SVD ($\lambda=4$) | 0.783 | 0.852 | 0.858 | 0.880 | 0.894 | 0.926 |
| LRTC (FaLRTC) | 0.588 | 0.685 | 0.702 | 0.746 | 0.777 | 0.834 |
| LRTC-STC ($p=1.7$, $u=10^6$) | 0.776 | 0.848 | 0.854 | 0.878 | 0.891 | 0.924 |
| Proposed algorithm | **0.564** | **0.660** | **0.671** | **0.709** | **0.728** | **0.787** |

Table 3: Imputation error on the HepB-II dataset with 40%-90% observed data

|  | 90% | 80% | 70% | 60% | 50% | 40% |
|---|---|---|---|---|---|---|
| meanImpute | 1.010 | 1.006 | 0.983 | 0.988 | 1.010 | 1.014 |
| Tensor-CP ($R=2$) | 0.858 | 0.865 | 0.833 | 0.832 | 0.861 | 0.881 |
| softImpute-ALS ($\lambda=4$) | 0.913 | 0.923 | 0.914 | 0.928 | 0.959 | 0.980 |
| softImpute-SVD | 0.914 | 0.924 | 0.914 | 0.929 | 0.963 | 0.980 |
| LRTC (HaLRTC) | 0.648 | 0.663 | 0.648 | 0.673 | 0.745 | 0.796 |
| LRTC-STC ($p=1.7$, $u=10^6$) | 0.918 | 0.927 | 0.915 | 0.929 | 0.961 | 0.976 |
| Proposed algorithm | **0.646** | **0.652** | **0.636** | **0.650** | **0.711** | **0.755** |